%% file: main.tex
\pdfoutput=1

\documentclass[11pt]{article}

\usepackage{acl}

\usepackage{times}
\usepackage{latexsym}
\usepackage{amssymb}
\usepackage{amsmath}
\usepackage{mathtools}
\usepackage{amsthm}
\usepackage{xcolor}
\usepackage{booktabs}
\usepackage{multirow}
\usepackage{multicol}
\usepackage[htt]{hyphenat}
\usepackage{balance}
\usepackage{bigfoot}

\DeclareNewFootnote{AAffil}[arabic]
\DeclareNewFootnote{ANote}[fnsymbol]

\usepackage[T1]{fontenc}

\usepackage[utf8]{inputenc}
\usepackage{inconsolata}

\usepackage{microtype}

\title{\method: Structured Open-Domain Dialogue Segmentation \\
and State Tracking in the Era of LLMs}

\author{
    Sarkar Snigdha Sarathi Das$^{1,\dagger,\ddagger}$, Chirag Shah$^{2,\ddagger}$,
    Mengting Wan$^{3}$, Jennifer Neville$^{3}$, \\ {\bf Longqi Yang$^{3}$, Reid Andersen$^{3}$, Georg Buscher$^{3}$, Tara Safavi$^{3,\dagger}$} \\ 
    $^1$Pennsylvania State University, $^2$University of Washington, $^3$Microsoft\\ 
    $^\dagger$Corresponding authors: \texttt{sfd5525@psu.edu}, \texttt{tarasafavi@microsoft.com} \\
    $^\ddagger$Work done at Microsoft, USA
}

\input{dfn}

\begin{document}
\maketitle
\begin{abstract}
\input{00abstract}
\end{abstract}

\section{Introduction}
\label{sec:intro}
\input{01intro}

\section{Problem Definition}
\label{sec:problem}

\input{02problem}

\section{Prompting Strategies}
\label{sec:prompt}
\input{03prompt}

\section{Experiments}
\label{sec:exp}
\input{04exp}

\section{Related Work}
\label{sec:related}
\input{05related}

\section{Discussion and Conclusion}
\label{sec:conclusion}
\input{06conclusion}


\bibliography{anthology}
\newpage

\appendix
\input{11appendix}
\balance

\end{document}

%% file: dfn.tex
\newcommand{\method}{S3-DST}

%% file: 00abstract.tex
The traditional Dialogue State Tracking (DST) problem aims to track user preferences and intents in user-agent conversations. While sufficient for task-oriented dialogue systems supporting narrow domain applications, the advent of Large Language Model (LLM)-based chat systems has introduced many real-world intricacies in open-domain dialogues. These intricacies manifest in the form of increased complexity in contextual interactions, extended dialogue sessions encompassing a diverse array of topics, and more frequent contextual shifts. To handle these intricacies arising from evolving LLM-based chat systems, we propose joint dialogue segmentation and state tracking per segment in open-domain dialogue systems. Assuming a zero-shot setting appropriate to a true open-domain dialogue system, we propose \method{}, a structured prompting technique that harnesses \textit{Pre-Analytical Recollection}, a novel grounding mechanism we designed for improving long context tracking. To demonstrate the efficacy of our proposed approach in joint segmentation and state tracking, we evaluate \method{} on a proprietary anonymized open-domain dialogue dataset, as well as publicly available DST and segmentation datasets. Across all datasets and settings, \method{} consistently outperforms the state-of-the-art, demonstrating its potency and robustness the next generation of LLM-based chat systems.

%% file: 01intro.tex
The advent of open-domain Large Language Model (LLM)-based chat systems like ChatGPT and Bing Chat has ushered in a new age of dialogue systems.
Previously, dialogue systems were relatively constrained in their scope and abilities, typically confined to either narrow task-oriented conversations or social chitchat~\cite{gao2018neural}.
By contrast, LLM-based chat systems are remarkable because they can converse fluidly with users over a seemingly infinite range of topics, and can accomplish many user tasks out-of-the-box that previously required specialized systems, like code generation, question answering, and more.  

\begin{figure}[t!]
     \centering
     \includegraphics[width=\columnwidth]{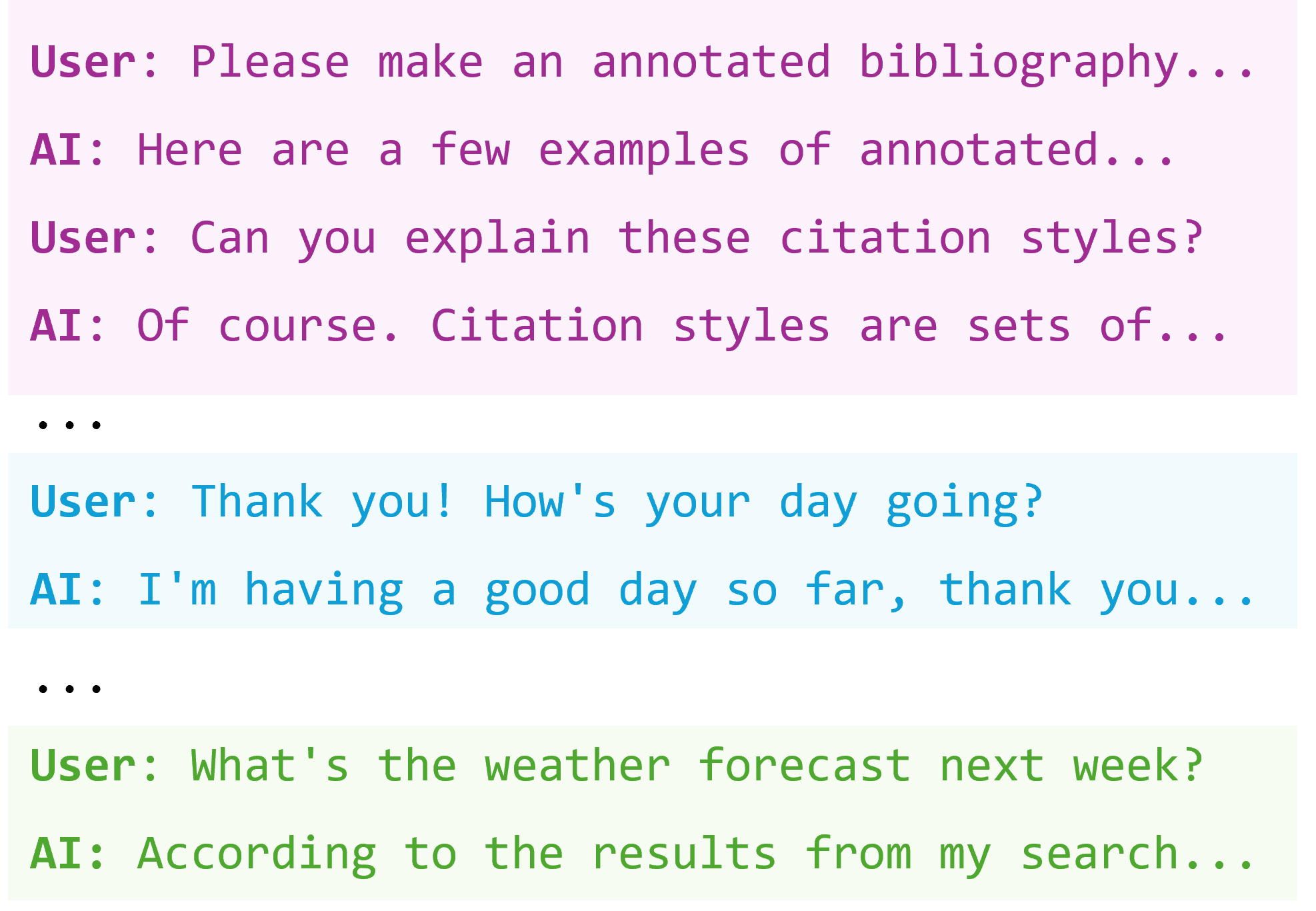}
    \caption{
    A single intent may span several turns in open-domain conversation, and a single conversation may contain multiple intents: A synthetic  dialogue inspired by anonymized Bing Chat logs. 
    Different user intents (creating an annotated bibliography, social chitchat, checking the weather) are highlighted by different colors. 
    }
    \label{fig:example}
\end{figure}

In this paper, we argue that because LLM-based chat systems have significantly changed the landscape of human-AI dialogue,  understanding user intent in such dialogues calls for new analysis and tagging frameworks. 
We focus in particular on the task of dialogue state tracking (\textbf{DST}).
Traditional DST consists of extracting and matching users' intents in task-oriented dialogue systems to a structured backend schema~\cite{williams2016dialog,budzianowski2018multiwoz}. 
However, DST in open-domain conversation is yet undefined; as such, in this paper we make a first attempt at identifying the state values of interest in LLM-based chat systems. 

As exemplified by Figure~\ref{fig:example}, we make the key observation that real open-domain dialogue often exhibits extensive back-and-forth between parties (e.g., clarification, negotiation, etc) in order to pursue a single intent or topic, and contexts may shift multiple times within a single dialogue among unrelated intents and/or topics.  
Based on this observation, we propose to track both \textbf{segments} and \textbf{states} in open-domain dialogue:
Segmentation helps us identify boundaries that mark the start and end of contextually cohesive conversation ``units,'' whereas states are the intent variables of interest we wish to track, applied \emph{per segment}. 

Beyond bringing DST into the era of open-domain conversation and LLMs, we introduce LLM-based \emph{solutions} for open-domain DST. 
Assuming a zero-shot setting for dialogue tagging, which is realistic due to the cost of labeling, we introduce \textbf{\method}, a structured prompting approach for open-domain DST. 
Within \method{} we propose a novel Pre-Analytical Recollection (PAR) prompting strategy that grounds each output state prediction on the content of the corresponding dialogue turn, thereby helping the LLM  track long dialogue context without forgetting or hallucination.

We evaluate \method{} on a fully anonymized open-domain dialogue dataset collected from Microsoft's Bing Chat system, alongside public DST and segmentation benchmarks.\footnote{The use of Bing Chat logs is in compliance with the terms of use of Bing Chat.}
\method{} achieves large gains over comparable baselines across all benchmarks, suggesting its suitability as a starting point for further research in open-domain dialogue modeling. 
In summary, our contributions are:
\begin{itemize}
    \item \textbf{Open-domain DST problem definition}: We bring dialogue state tracking into the era of open-domain LLM chat.
    We cast the problem as a joint segmentation and state tracking task, motivated by our observations of how real open-domain human-AI conversation is conducted on anonymized Bing Chat log data.
    \item \textbf{Zero-shot \method{} approach}:
    We propose \method{}, a \textbf{s}tructured zero-shot joint \textbf{s}egmentation and \textbf{s}tate tracking approach for open-domain, multi-intent dialogue. 
    \method{} contributes new approaches for structured prompt templating and dialogue tag generation, as well as Pre-Analytical Recollection (PAR), a grounding technique that improves long context tracking. 
    \item \textbf{Extensive experiments and analysis}:
    We conduct extensive experiments on both proprietary and public datasets, achieving large gains over comparable zero-shot prompts.
    \method{} achieves state-of-the-art zero-shot performance on the MWOZ 2.1 and 2.4 DST benchmarks, alongside the DialSeg711 dialogue topic segmentation benchmark.
\end{itemize}

%% file: 02problem.tex
Informally, the goal of traditional DST is to predict the dialogue state $y_t$ given a sequence of user and agent utterance turns $C_t = [ U_1, A_1, \hdots, U_t, A_t ]$.\footnote{Note that in current LLM-based chat systems, users may issue multiple utterances before a single agent response is issued. In these (infrequent) cases, we group all user utterances prior to the agent response into a single utterance. }
The state $y_t$ consists of a set of slot-value pairs, where slots correspond to intent attributes in a particular application domain (e.g., ``restaurant-name'', ``hotel-address'') and values correspond to predefined categorical options or unconstrained text~\cite{budzianowski2018multiwoz}.

However, as we have previously discussed, a single open-domain conversation will often consist of multiple potentially unrelated intents across a variety of topics.
Indeed, 
according to a preliminary analysis on 10K anonymized Bing Chat conversations, we estimate that over 50\% of conversations display multiple user intents and over 90\% of conversations contain discussion of multiple topics. 
Therefore, we propose to merge dialogue segmentation, which aims to find contextually cohesive ``units'' of dialogue within a larger conversation, with dialogue state tracking.
In particular, we perform state tracking at the \emph{segment} level, where the goal is to label each segment with the slots and values of interest, such that multiple segments within a conversation may have diverging or conflicting  state values, reflecting the true variety of open-domain chat. 

In the rest of this section, we define segmentation and state, and finally formalize the joint task. 

\subsection{Segment}
\label{problem:segment}

Following previous work in dialogue topic segmentation~\cite{xing-carenini-2021-improving,xia2022dialogue,gao2023unsupervised}, we define \textbf{dialogue segments} as contiguous subsequences of $C_t$ in which all user and agent utterances are topically related. 
Formally, let $B_t = [b_1, \hdots, b_{t-1}]$ indicate the boundary indices between adjacent user-agent utterance pairs in $C_t$.
The output of segmentation is a set of boundary indices $B_{k} \subseteq B_t$, where $k$ represents the number of boundaries determined by the segmentation algorithm and the span $[U_{m}, A_{m}, \hdots U_{n}, A_{n}]$ represents the contiguous segment  between boundaries $b_m$ and $b_{n}$, where $m \in [1, t - 1]$ and $n \in [m, t - 1]$. 

\subsection{Segment state}
\label{problem:state}

Typically, dialogue state tracking methods extract new elements of state at each turn~\cite{hu-etal-2022-context}.
However, this is because DST evaluation benchmarks make the relatively narrow assumption that users provide new and relevant elements of intent at each turn, and that intents build upon or complement each other but do not fundamentally change or conflict throughout the conversation. 
As we have previously discussed, open-domain dialogue exhibits far more varied characteristics, and multi-intent and/or multi-domain conversations are relatively common. 

We therefore propose to extract state at the segment rather than turn level. 
We define the segment-level state as 
$\{S_{m:n} = (s_{m:n}^{(i)}, v_{m:n}^{(i)}), i = 1 \hdots N_{m:n}\}$, 
where $s_{m:n}^{(i)}$ refers to the $i$-th slot applied to the segment from boundaries $b_m$ to $b_{n}$, $v_{m:n}^{(i)}$ refers to the slot's corresponding value, and $N_{m:n}$ refers to the total number of slots to applied to this segment. 
Any schema of slot-value pairs is valid here; we describe our particular state schema for Bing Chat in \S~\ref{data:bing-chat}  and Appendix~\ref{app:annotation}. 

\subsection{Problem statement}
\label{problem:def}

Having defined segments and per-segment state, we are equipped to state our full definition of open-domain DST. 
Given a sequence of user-agent utterance pairs $C_t = [ U_1, A_1, \hdots, U_t, A_t ]$, we define the goal of open-domain dialogue state tracking as jointly predicting
\begin{align}
\label{prob-def}
y_t &= B_k \cup \{ S_{m:n} \, ; \, \forall (b_m, b_n) \in B_k \},
\end{align}
where $B_k \subseteq B_t$ refers to the segment boundary indices described earlier and $S_{m:n}$ refers to the segment state between boundaries $b_m$ and $b_n$, consisting of $N$ arbitrary slot-value pairs: 
\begin{align}
S_{m:n} &= \{(s_{m:n}^{(i)}, v_{m:n}^{(i)}), i = 1 \hdots N_{m:n}\}. 
\end{align}

\begin{figure*}[h]
     \centering
    \includegraphics[width=\textwidth]{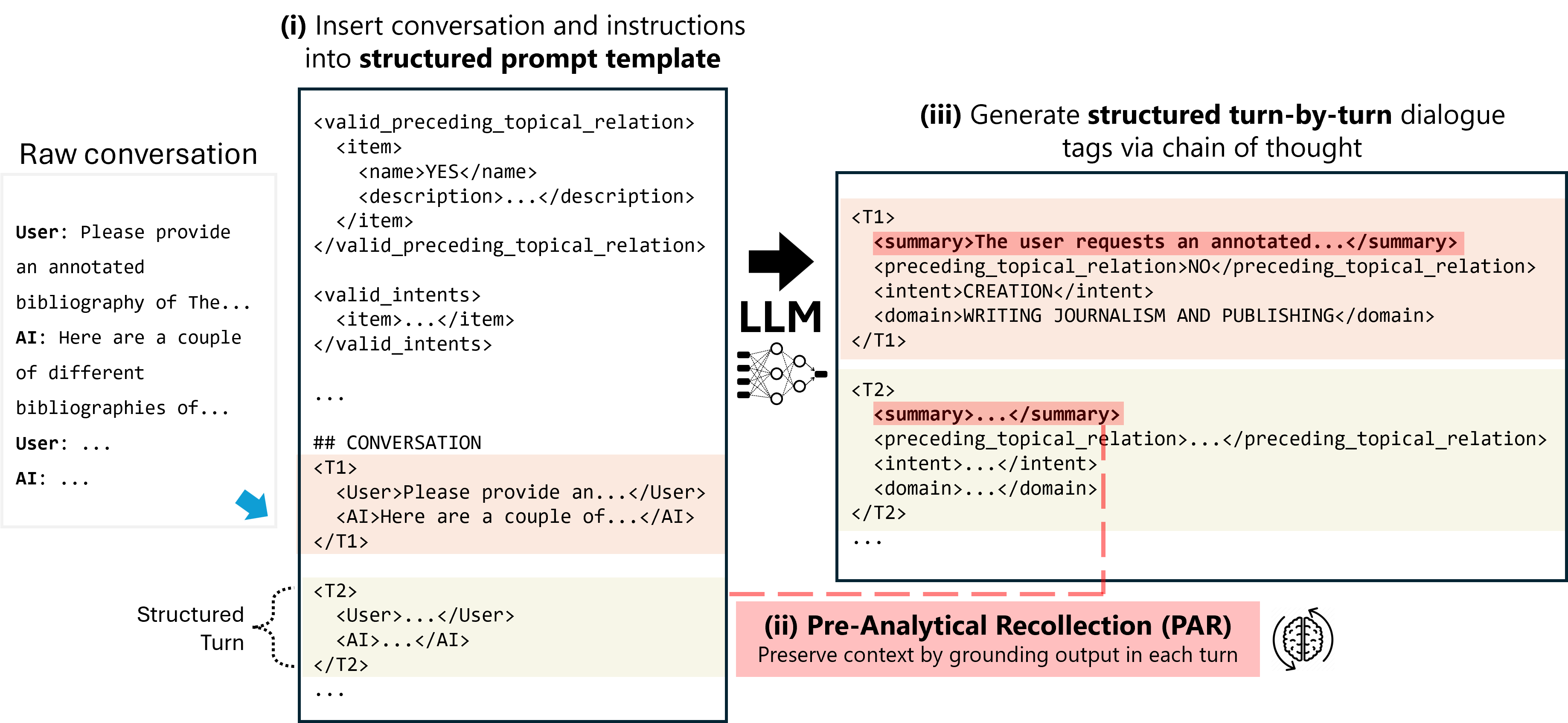}
    \caption{
     Prompt flow of \method. Given a raw conversation, (i) we convert it into a hierarchical XML-structured representation and insert it into a similarly structured prompt template.
     We pass the prompt through the LLM and (ii) obtain a hierarchical XML-structured output, where each turn contains (iii) a PAR grounding reference to the conversation alongside the desired segmentation and state label predictions. 
    }
    \label{fig:s3-dst}
\end{figure*}

%% file: 03prompt.tex
As discussed previously, real-world dialogues often exhibit extensive discourse that extends over multiple conversational turns in order to discuss diverse topics. This prolonged conversational nature makes it highly challenging to track contextual coherence. Previous studies \cite{hu-etal-2022-context} aimed at disassociating individual dialogue turns and processing them one by one for tracking dialogue state changes, which worked reasonably well in task-oriented dialogues confined within predefined narrow domains. 

However, real-world dialogues commonly require multiple turns to adequately comprehend the contextual nuances, which is a challenge because Transformers still struggle when processing lengthy input contexts, particularly in the middle \cite{liu2023lost}.
To address these difficulties, we propose a novel turn-by-turn prompting technique that gives structure to inputs and outputs while accurately preserving the context in the process. We discuss these design aspects of our prompts below: 

\subsection{Structured Outputs and Inputs}
\label{method:xml}

\paragraph{Structured Output} 
Our goal is a set of labels per dialogue turn representing the segment boundaries (binary labels) and state values (categorical labels or open text). 
To provide a flexible yet structured format to the LLM's output, we propose to instruct it to generate outputs in a hierarchical XML format.
We see XML as advantageous because it provides code-like structure to the DST task, which has been shown to greatly improve performance compared to plain-text outputs, while still being extensible and flexible compared to more rigid output formats like SQL~\cite{hu-etal-2022-context}. 

Our approach uses an XML format in which each turn from 1 to $t$ comprises an XML tree \texttt{<T\{id\}>...</T\{id\}>} and several nested XML tags within it. 
The labels of these nested tags (e.g. \texttt{<preceding\_topical\_relation>...</preceding\_topical\_relation>}, \texttt{<intent>...</intent>}, and \texttt{<domain>...</domain>}  in Figure~\ref{fig:s3-dst}(iii)) represent  the segment boundaries and slots of interest, and each value between opening and closing tags represent the model's inferred value. 

This strategy is beneficial from two fronts: (i) Due to bounded well-defined structured formatting, generated outputs are more likely to be aligned with labeling instructions than free-form texts, and (ii) Well-formed structured output formats are easier to parse, thus reducing postprocessing requirements.

\paragraph{Structured Input} For prompting LLMs, although it is trivial to channel plain conversation history in a flat format for analysis and inference, the unstructured nature inherent to this linear configuration makes it difficult to refer back and leverage different information across multiple conversational turns. To handle this challenge, consistent with the output format, we propose a structured inputting format, where each conversational history is formed into a hierarchical XML format where conversational turns are marked with turn id number \texttt{<T\{id\}>...</T\{id\}>} numbered from $1$ to $t$ and each conversational turn consists of nested user and agent turns marked with appropriate XML tags (\texttt{<user>...</user>} and \texttt{<agent>...</agent>}). 

Since we propose instructing the LLM to infer per-turn labels during our output, this input scheme helps us accurately refer back to the input turn and thus maintain coherence even for long dialogue contexts. Consistent with this XML-tagged input format, we also format all the valid segment and state categories in an XML-formatted list using the following structure:
\texttt{<valid\_category\_name>} \texttt{<item>}\{label name\}\texttt{</item>} 
\texttt{<description>}
\{description of label, if available\}
\texttt{</description>} \texttt{<valid\_category\_name>}
Empirically, this structured input and prompt formatting help constrain the LLM generation to follow the labeling instructions. Figure \ref{fig:s3-dst}(i) shows this format where each valid segment boundary and state category are first staged in an XML-formatted list and subsequently input dialogue is shown in a hierarchical configuration.

\subsection{Pre-Analytical Recollection (PAR)}
\label{method:par}

As previously discussed, open-domain dialogues may be long and highly variable in conversation flow.
Therefore, it is crucial to ensure that the LLM can accurately monitor the evolving dialogue context without forgetting or hallucination. 
To this end, we propose Pre-Analytical Recollection (PAR), a grounding strategy for turn-by-turn prompting that instructs the LLM to first summarize the turn using \texttt{<summary>...</summary>} tags in 3 sentences or fewer before providing the segment and state values. 
PAR is inspired by chain-of-thought prompting~\cite{wei2022chain}, as it is a technique for generating relevant intermediary outputs in order to improve reasoning accuracy. 
However, unlike chain-of-thought, PAR is also a grounding technique that provides references from the model's output directly to the conversation. Figure \ref{fig:s3-dst}(ii) demonstrates how PAR refers back to the content of each conversational turn before analyzing it to infer the conversational states.

\subsection{Final Prompt Configuration}
\label{method:prompt}

The final prompt flow of \method{} is provided in Figure~\ref{fig:s3-dst}. 
Given a raw conversation and a predefined set of segment and state labels,
we insert the labels into a structured prompt template and format the conversation in a hierarchical XML-structured representation.
We pass the prompt through the LLM, instructing it to follow PAR before jointly generating the hierarchical turn-by-turn segmentation and state labels applied per segment. 
The full text of our prompt is provided in Appendix~\ref{app:our-prompts}. 

%% file: 04exp.tex
\begin{table}[t!] 
\centering 
\caption{Evaluation test set statistics.} 
\label{table:data}
\resizebox{0.99\columnwidth}{!}{
\begin{tabular}{l rrr} 
\toprule  
& \# Convs & \# Turns & \# segments/conv  \\ 
& & & (avg.) \\ 
\midrule 
Bing Chat & 334 & 2308 & 1.51 \\ 
MWOZ 2.1 & 1,000 & 7368 & - \\ 
MWOZ 2.4 & 1,000 & 7368 &  - \\ 
DialSeg711 & 711 & 19350 & 3.87 \\ 
\bottomrule 
\end{tabular}
}
\end{table}

We conduct comprehensive evaluations across multiple datasets. We primarily evaluate our approach on fully anonymized Bing Chat logs annotated by domain experts. Additionally, we evaluate \method{} on the standard task-oriented DST and segmentation tasks using public benchmark datasets MultiWOZ \cite{budzianowski2018multiwoz} and DialSeg711 \cite{xu2021topic} respectively. A detailed description of these datasets is provided below, alongside dataset statistics in Table~\ref{table:data}:

\subsection{Internal Human-LLM Dialogue Dataset}
\label{data:bing-chat}

In order to evaluate the efficacy of our approach on real-world open-domain human-LLM conversations, we collected anonymized chat log data from Microsoft's Bing Chat system, an LLM chat interface backed by the Bing search engine.
\paragraph{Benchmark construction}
We sample 484 English conversations conducted on Bing Chat between April 5, 2023 to April 30, 2023 via two approaches: (i) Random and (ii) ``Long'' conversations of 5 or more turns only. 
We balance these two approaches 50/50. 
Since we operate under a zero-shot assumption, we do not need any training data.
Therefore, we hold out 150 conversations for development and the remaining 334 for testing. 

\paragraph{Annotation}
To obtain ground-truth labels for evaluation, we gathered human annotations for segment and state.  
We recruited three in-house annotators with a high degree of technical expertise and familiarity with the Bing Chat system.

For each turn, we instructed annotators to provide binary \textbf{IsSegmentBoundary} labels, categorical \textbf{SegmentIntent} labels, and categorical \textbf{SegmentDomain} labels.
We instructed annotators to mark a segment boundary when no topical relation between a turn and its preceding context could be identified. 
For intent and domain, we used taxonomies developed in-house for the Bing Chat system consisting of 4 intents (Information Seeking, Analysis, Creation, and Open-Ended Discovery) and 49 domains (see Appendix~\ref{app:annotation-labels} for the full list). 
Because of the large number of domains, per turn we provided annotators four candidate domain values and an ``Other'' option.
Appendix~\ref{app:annotation} provides further details on the annotation scheme and domain sampling procedure. 
To ensure inter-annotator agreement before labeling the full dataset, we first gathered annotations on a set of 10 randomly selected conversations (68 turns total) and computed Fleiss' kappa~\cite{fleiss1971measuring} per label type. 
We observed a Fleiss kappa of $\kappa=0.83$ for \textbf{IsSegmentBoundary}, $\kappa=0.74$ for \textbf{SegmentIntent}, and $\kappa=0.88$ for \textbf{SegmentDomain}, all of which are considered high agreement on the Fleiss kappa scale. 

\subsection{Public Benchmarks}
\label{data:public}

We are not aware of any existing public dialogue benchmarks reflective of the broadly open-domain Bing Chat data. 
Therefore, we resort to separate DST and segmentation evaluations on public benchmarks using three datasets.

\paragraph{MultiWOZ} 
The MultiWOZ (MWOZ) multi-domain dialogue dataset~\cite{budzianowski2018multiwoz} is currently the most common DST benchmark.
MWOZ is a task-oriented dataset consisting of 1K test dialogues. 
We use two updated versions of the original: MWOZ 2.1~\cite{eric2019multiwoz21} and 2.4~\cite{ye2021multiwoz24}.
The latter is considered the ``cleanest'' version of MWOZ, while the former has been used more frequently in the literature. 

\paragraph{DialSeg711}
The DialSeg711 benchmark was introduced by~\cite{xu2021topic} and has been used frequently in recent dialogue segmentation research. 
It is an English dataset in which 711 multi-segment dialogues are constructed by joining dialogues from existing task-oriented dialogue corpora.

\begin{table*}[t!] 
\centering 
\caption{\method{} achieves state-of-the-art performance on state tracking over our internal Bing Chat benchmark.
All prompts are run with GPT4. 
} 
\label{table:bingchat}
\resizebox{0.8\textwidth}{!}{
\begin{tabular}{l ccc c cc} 
\toprule 
& \multicolumn{3}{c}{Individual accuracy} && \multicolumn{2}{c}{JGA} \\ 
\cline{2-4} \cline{6-7}
& Segment & Intent & Domain && I/D & S/I/D \\ 
\midrule 
TBT-DST & - & 0.6707 & 0.6221 && 0.4169 & - \\ 
IC-DST & 0.8567 & 0.7123 & 0.6049  && 0.4610 & 0.4387 \\ 
\method{} (No PAR) & 0.8859 & 0.7173 & 0.6251 && 0.4377 & 0.4078 \\ 
\method{} (Unstructured input) & 0.8810 & 0.7163 & 0.6307 && 0.4640 & 0.4331 \\ 
\midrule 
\textbf{\method{}} & \textbf{0.8992} & \textbf{0.7366} & \textbf{0.6429} && \textbf{0.4752} & \textbf{0.4504} \\ 
\bottomrule 
\end{tabular}
}
\end{table*}

\subsection{Baselines}
\label{exp:baselines}

As baselines we consider zero-shot LLM prompts only, for a fair comparison to \method.
We discuss the baselines and their considerations below for different datasets.
All original prompts are provided in Appendix~\ref{app:prompts}. 
We set a maximum of 1500 output tokens per LLM call with a temperature of zero. 

\paragraph{Bing Chat} 
In this dataset, we consider \textbf{IC-DST} as our primary baseline, which is a zero-shot version of the prompting strategy introduced by \cite{hu-etal-2022-context}, heavily adapted for open-domain dialogue setting to jointly track segment and dialogue states. The \textbf{TBT-DST} baseline is a version of \method{} that does not include segmentation instructions and obtains intent and domain labels on a turn-by-turn basis using our \method{} prompt configuration. Moreover, to analyze the importance of two key aspects of our prompt, PAR and XML-structured formatting,  we also consider two ablations of \method: \textbf{No PAR} refers to a \method{} prompt without the PAR instructions, and \textbf{Unstructured input} refers to a \method{} prompt that formats all instructions and dialogue using plain text rather than XML.  
We use GPT4 as the backbone LLM for all prompts. 

\paragraph{MWOZ}
For MWOZ task-oriented dialogue state tracking dataset, we compare against \textbf{IC-DST using Codex-175B} as reported by~\citet{hu-etal-2022-context}. We also reevaluate zero-shot \textbf{IC-DST with GPT-4} to account for the backbone model improvement in baseline performance. Finally, we compare against the zero-shot \textbf{ChatGPT} performance on MWOZ 2.1 as reported by \cite{heck-etal-2023-chatgpt}.

\paragraph{DialSeg711}
We consider the unsupervised \textbf{TextTiling}~\cite{hearst1997text}, \textbf{CSM}~\cite{xing-carenini-2021-improving}, and \textbf{DialStart}~\cite{gao2023unsupervised} methods. 
We reprint all numbers from~\cite{gao2023unsupervised}. 
Finally, we use our \textbf{IC-DST} baseline prompted to elicit segmentation labels in the same SQL output format as the original IC-DST~\cite{hu-etal-2022-context}.

\subsection{Metrics}
\label{exp:metrics}

For state tracking, we consider \textbf{Joint Goal Accuracy (JGA)}, which measures the proportion of turns for which all state values are correctly inferred. 
For Bing Chat, we report JGA with just intent and domain (I/D) as these are the true state values of interest, as well as JGA with segment, intent, and domain accuracy (S/I/D) for completeness. 
We also report segmentation, intent, and domain accuracy separately on Bing Chat to provide a sense of the current capabilities and limitations of LLMs on open-domain conversational data.
For segmentation, we consider $P_K$ and \textbf{WindowDiff}~\cite{pevzner2002critique}, which are both error metrics (i.e., lower is better) that quantify the difference between predicted and ground-truth segment boundaries using an adjustable sliding window. 

\begin{figure}[t!]
     \centering
    \includegraphics[width=\columnwidth]{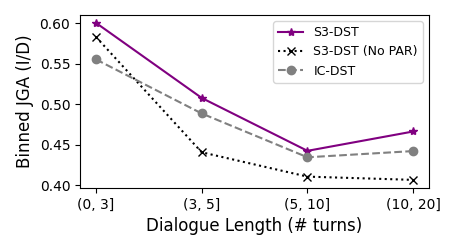}
    \caption{
     \method{} outperforms baselines for dialogues of all lengths by emphasizing context tracking.  
     We bin Bing Chat dialogues by length and plot JGA per bin.
     The large performance degradation of both baselines as the dialogue length increases confirms the importance of our PAR grounding strategy. 
    }
    \label{fig:jga-vs-length}
\end{figure}

\begin{table}[t!] 
\centering 
\caption{\method{} achieves state-of-the-art JGA compared to zero-shot LLM baselines on the public dialogue state tracking benchmarks MWoZ 2.1 + 2.4.} 
\label{table:mwoz}
\resizebox{0.9\columnwidth}{!}{
\begin{tabular}{lcc} 
\toprule 
& \multicolumn{2}{c}{JGA} \\ 
\cline{2-3}
& MWOZ 2.1 & MWOZ 2.4 \\ 
\midrule 
IC-DST (Codex) & 0.3534 & 0.3530 \\
IC-DST (GPT4) & 0.4045 & 0.4625 \\
ChatGPT & 0.3150 & - \\ 
\midrule 
\textbf{\method{}} & \textbf{0.4513} & \textbf{0.5327} \\ 
\bottomrule 
\end{tabular}
}
\end{table} 

\begin{table}[t!] 
\centering 
\caption{Zero-shot per-domain comparison (JGA) on MWOZ 2.1.} 
\label{table:mwoz-per-domain}
\resizebox{\columnwidth}{!}{
\begin{tabular}{l ccccc} 
\toprule 
& \multicolumn{5}{c}{Per-domain JGA} \\ 
\cline{2-6}
& attr. & hotel & rest. & taxi & train \\ 
\midrule 
IC-DST (Codex) & 0.5997 & 0.4669 & 0.5728 & 0.7135 & 0.4937 \\ 
IC-DST (GPT4) & \textbf{0.7177} & 0.4872 & 0.6526 & 0.7781 & 0.5710 \\ 
ChatGPT & 0.5270 & 0.4200 & 0.5580 & 0.7090 & 0.6080 \\ 
\midrule 
\textbf{\method{}} & 0.6781 & \textbf{0.5215} & \textbf{0.6713} & \textbf{0.8258} & \textbf{0.7027} \\ 
\bottomrule 
\end{tabular}
}
\end{table} 

\subsection{Results}
\label{exp:results}

\paragraph{Bing Chat}
As shown in Table~\ref{table:bingchat}, our \method{} prompt achieves the highest performance across intent, domain, and JGA across turns. 
We make the following observations:
First, TBT-DST, which does not explicitly perform segmentation, is by far our weakest baseline. 
We find that this is because without instructing the LLM to use the same intent and domain within a segment, the LLM tends to overindex on the content of the turn without considering the fuller preceding context.
This leads to conflicting intent and domain labels between turns within a coherent single-topic dialogue. 
    
Second, our adapted version of IC-DST is a very strong baseline.
However, while IC-DST makes use of structured outputs, it does not have a corresponding structured input representation.
We find that this hurts its performance in some cases, as hallucination of nonexistent turns is relatively more common compared to \method. 

Finally, the two ablations of \method{} both underperform compared to \method, confirming the importance of PAR and structured inputs that the LLM can refer back to during generation. 
Indeed, Figure~\ref{fig:jga-vs-length}, which plots the relationship between dialogue length and performance, shows that \method{} avoids the steep degradation in performance of the no-PAR ablation as the dialogues get longer.
For example, the no-PAR ablation performs comparably to \method{} on conversations of 3 turns or fewer, but drops over 10 points JGA for conversations of 4 turns or more. 
These results in particular highlight the necessity of PAR for long dialogues.

\paragraph{MWOZ}
Tables~\ref{table:mwoz} and~\ref{table:mwoz-per-domain} provide MWOZ numbers in total and per-domain.
\method{} achieves state-of-the-art zero-shot JGA compared to strong LLMs by a large margin.
Even our strongest zero-shot baseline, IC-DST (GPT4), has an absolute performance gap of nearly 5 points JGA on MWOZ 2.1 and 7 points on MWOZ 2.4. 
In nearly all individual domains, \method{} outperforms IC-DST (GPT4), and some by a large margin, for example over 13 points JGA improvement on the train domain. 

\begin{table}[t!] 
\centering 
\caption{\method{} achieves state-of-the-art performance on the public segmentation benchmark DialSeg711.} 
\label{table:dialseg}
\resizebox{0.7\columnwidth}{!}{
\begin{tabular}{l rrrrr} 
\toprule 
& $P_k (\downarrow)$ & WindowDiff $(\downarrow)$  \\ 
\midrule 
TextTiling & 0.4044 & 0.4463 \\ 
CSM & 0.2430 & 0.2635  \\ 
DialSTART & 0.1786 & 0.1980 \\ 
IC-DST & 0.2889 & 0.2419 \\ 
\midrule 
\textbf{\method{}} & \textbf{0.0091} & \textbf{0.0081} \\ 
\bottomrule 
\end{tabular}
}
\end{table} 

\paragraph{DialSeg711}
Finally, Table~\ref{table:dialseg} shows performance on DialSeg711.
\method{} achieves nearly zero error on this dataset, which we find unsurprising given that the dataset's construction.
Specifically, DialSeg711 is constructed by joining dialogues about very different topics, which leads to very artificial and abrupt context shifts between segments.
However, we find that our IC-DST prompting baseline leads to much higher error than \method. 
On further inspection, we find that the LLM fails to track the dialogue context for several conversations in the dataset, leading to forgetting of the original conversation context.
These results highlight the importance of PAR and dialogue context tracking for successful segmentation. 
\method{}'s strong performance also suggests that DialSeg711  may not be a difficult enough task in future for LLMs, and further motivates the need for joint segmentation and state tracking, as the goal of segmentation is ultimately to improve state tracking performance. 

%% file: 05related.tex
\subsection{Dialogue State Tracking}
To accurately track the passage of Human-AI conversation, robust state tracking is crucial toward inferring user intentions and goals. Since the introduction of the MultiWOZ \cite{budzianowski2018multiwoz} dataset to the community, a plethora of techniques have been proposed to improve DST performance. Earlier attempts including copy mechanism \cite{lei2018sequicity}, transfer learning \cite{wu-etal-2019-transferable}, data augmentation \cite{zhang2020task}, contrastive pretraining \cite{wu-etal-2020-tod}, etc. have yielded improvements in supervised fine-tuning scenarios; meanwhile, MultiWOZ also went through several annotation revisions \cite{eric2019multiwoz21, ye2021multiwoz24, zang2020multiwoz, han2020multiwoz}. While other techniques \cite{peng2021soloist,lin2020mintl, zhao2022description, yu2020score, platanios-etal-2021-value} have also been proposed, the resource-intensive and laborious nature of data labeling has gradually redirected attention toward the exploration of few- and zero-shot dialogue state tracking \cite{shin-etal-2022-dialogue, hu-etal-2022-context, heck-etal-2023-chatgpt}. While the state-of-the-art approach in this discipline \cite{hu-etal-2022-context} can leverage LLMs for tracking states, it notably lacks proper grounding mechanisms which can potentially hurt performance in real-world extended dialogue sessions. Furthermore, none of the aforementioned previous work accounts for topic coherence and context switches prevalent in flexible open-domain LLM-based chat systems.

\subsection{Dialogue Topic Segmentation}

Segmenting a dialogue into topically coherent units is foundational to successful downstream dialogue modeling. While the paucity of annotated data has been a challenge in dialogue topic segmentation, recent unsupervised attempts have exhibited some promising outcomes in topic segmentation. More specifically, extensions based on the classical text segmentation algorithm TextTiling \cite{hearst1997text} have primarily led the benchmark in this aspect~\cite{song2016dialogue}. More recently, text-pair coherence scoring \cite{xing-carenini-2021-improving} and topic-aware representation learning \cite{gao2023unsupervised} have advanced the state of the art. Nevertheless, all of these techniques fall short in accounting for the complete contextual essence of a conversation (i.e., explicitly modeling intent and other important state variables), which can lead to suboptimal results.

\subsection{Intent Classification}
Related to dialogue state tracking, another fundamental problem in task-oriented dialogue systems is intent classification (IC). Often paired with another complementary problem slot-filling (SF), researchers have proposed a wide range of techniques over the years \cite{liu2016attention,zhang2016joint,goo2018slot, qin2019stack,qin2021co}, achieving impressive performance in popular public datasets. Few-shot techniques have also been investigated in data-constrained scenarios for joint IC/SF task \cite{krone2020learning, bhathiya2020meta, liu2021explicit}. While related to DST, IC/SF primarily deals with individual utterances in isolation, which makes it less apt for real-world human-AI dialogue which often requires modeling intricate contextual connections spanning multiple utterances within a conversational session. 

%% file: 06conclusion.tex
LLM-based chat systems have broadened the horizons  of human-AI conversation, warranting new methods for tracking user intentions. Therefore, we bring dialogue state tracking in the realm of open-domain dialogue systems by jointly tracking topically coherent segments and state intent variables per segment. Since this requires the assumption of a zero-shot setting due to the impracticality of annotation across all disciplines, we propose \method{}, a structured segmentation and state tracking approach using zero-shot prompting for open-domain state tracking. \method{} structures the prompt in an XML format and leverages our proposed grounding mechanism (PAR) for long context tracking. Across extensive experiments on proprietary and public datasets, \method{} shows large performance gains over state-of-the-art zero-shot techniques in dialogue state tracking and segmentation approaches. 
In the future, as LLM-based chat systems become more prevalent, we expect dialogue systems research to shift further toward understanding and modeling open-domain dialogue.
In this respect, we aim to further study and develop techniques for extended context preservation, to improve grounding in DST alongside other important dialogue modeling tasks.

%% file: 11appendix.tex
\section{Prompts}
\label{app:prompts}

\subsection{\method{} prompts}
\label{app:our-prompts}

\paragraph{Bing Chat}
Below is the full prompt for \method{}, with templated values to be replaced by e.g., intent label names or descriptions in curly braces. Appendix~\ref{app:annotation} provides the full list of state values. 

\noindent\texttt{\small{<valid\_domains> \\
<item>\{\textrm{valid domain label name}\}</item>  \\
\textrm{...} \\ 
</valid\_domains> \\
<valid\_preceding\_topical\_relation> \\ 
<item> \\ 
<name>YES</name> \\
<desc>The current turn has **some or any** topical/subtopical relation to the preceding conversation context.</desc>
</item> \\
<item> \\ 
<name>NO</name> \\
<desc>The current turn has **absolutely no** topical/subtopical relation to the preceding conversation context OR is the first turn in the conversation, marking the beginning of a new dialogue segment.
</desc> \\
</item> \\
</valid\_preceding\_topical\_relation> \\
<valid\_intents> \\ 
<item> \\
<name>\{\textrm{valid intent label name}\}</name> \\
<desc>\{\textrm{intent description}\}</desc> \\
</item> \\ 
\textrm{...} \\ 
</valid\_intents> \\
\#\# TASK \#\# \\ 
You are given a dialogue between a user and an agent comprised of turns starting with T. For each turn you have to answer the following questions.  \\
- Summarize the turn in <=3 sentences \\
- Output the preceding\_topical\_relation label using the <valid\_preceding\_topical\_relation>...</valid\_preceding\_topical\_relation> list \\
- Output the intent label from the <valid\_intents>...</valid\_intents> list \\
- Output the domain label from the <valid\_domains>...</valid\_domains> list \\
- When preceding\_topical\_relation is YES, you must use the exact same intent and domain label for all turns in the segment. \\
\#\# OUTPUT FORMAT \#\# \\
<T\{turn number\}> \\
<summary>\{turn summary in <=3 sentences\}</summary> \\
<preceding\_topical\_relation>\{valid preceding topical relation label\}</preceding\_topical\_relation> \\
<intent>\{valid intent label\}</intent> \\
<domain>\{valid domain label\}</domain> \\
</T\{turn number\}> \\
\#\# INPUT \#\# \\
\{\textrm{XML-structured dialogue\}} \\
\#\# OUTPUT \#\# \\
}
}

For the ``No PAR'' baseline, we remove the turn summarization instruction and summary tag from the prompt.
For the ``Unstructured input'' baseline, we input the conversation as a list of plain-text turns numbered from T1 to T\{$t$\}.
For the TBT-DST baseline, we remove all segmentation instructions and labels from the prompt, and simply have the model output a valid intent and domain per turn.

For the DialSeg711 dataset, we remove all instructions and values related to intent and domain, and have the model output turn-level summaries and segment labels only. 

\paragraph{MWOZ}
Below is the \method{} prompt for the MWOZ dataset.
Note that all descriptions for slots were generated by GPT4. 

\noindent\texttt{\small{<slots> \\ 
<item> \\ 
<name>taxi-leave at</name> \\
<description>the time when the user wants to get the taxi</description> \\ 
</item> \\ 
<item> \\ 
<name>\{\textrm{domain}\}-\{\textrm{intent}\}</name> \\ 
<description\{\textrm{description of slot}\}</description> \\ 
<valid\_values>\{\textrm{valid categorical values for slot if applicable, otherwise this tag does not appear}\}</valid\_values> \\
</item>  \\ 
\textrm{...}  \\ 
</slots> \\ 
\#\# TASK \#\#  \\ 
You are given a dialogue between a user and an agent comprised of turns starting with T. For each turn you have to answer the following questions. \\ 
- Output the user utterance verbatim. \\ 
- Based on that utterance, extract the relevant information about user preferences for relevant slots from <slots>...</slots> and represent them as a list of tags that follow the format ['\{SLOT\}-\{value\}'], where value is the specific information for that SLOT. \\ 
- Remove any duplicates or conflicting pairs from the list. If the same SLOT appears more than once in the list, keep only the most recent or relevant value originated from a user utterance. \\ 
- If the values for the same SLOT contradict each other, resolve the conflict by keeping the **most recent** user provided value. Output the final list as the task result. \\ 
- Example output for ['\{SLOT\}-\{value\}']. For example, the output may look like ['hotel-book day-monday', 'hotel-book number\_of\_people-3', 'hotel-book number\_of\_days-4', 'hotel-name-wartworth', 'hotel-area-east', 'hotel-parking-yes', 'hotel-stars-4', 'hotel-internet-yes', 'train-book number\_of\_people-1', 'train-destination-bishops stortford', 'train-day-friday', 'train-arrive\_by\_time-19:45', 'train-departure-cambridge'] \\ 
- Make sure selected slots are only from predefined <slots>...</slots> list. If <valid\_values>...</valid\_values> are mentioned for the slot, you must use one of the valid values for that slot. \\ 
- Use dontcare values only if user explicitly mentions it. \\ 
Now for **every turn**, answer the following questions:  \\ 
<T\{turn number\}> \\ 
<agent\_context> \{verbatim last agent utterance\} </agent\_context> \\ 
<user\_utterance> \{verbatim user utterance of the turn\} </user\_utterance> \\ 
<updated\_slot\_value> updated list of ['\{SLOT\}-\{value\}'] taking slots from <slots>...</slots> and using <valid\_values>...</valid\_values> for appropriate slots </updated\_slot\_value>
</T\{turn number\}> \\ 
\#\#INPUT\#\#  \\ 
\{\textrm{XML-structured dialogue}\} \\
\#\#OUTPUT\#\#  \\ 
}
}

\subsection{IC-DST prompt}
\label{app:baseline-prompts}

Below is the IC-DST prompt adapted to the Bing Chat dataset.
Note that for the DialSeg711 dataset, we simply remove the domain and intent columns and instructions.

\noindent\texttt{\small{CREATE TABLE states( \\ 
domain text CHECK (domain IN (\{\textrm{valid domain names})), \\
preceding\_topical\_relation text CHECK (preceding\_topical\_relation IN (YES, NO)), \\
intent text CHECK (intent IN (\{\textrm{valid intent names})), \\
) \\ 
/* \\ 
\#\# DESCRIPTION OF SELECTED COLUMN-VALUE PAIRS: \\
- preceding\_topical\_relation-NO: The current turn has **absolutely no** topical/subtopical relation to the preceding conversation context OR is the first turn in the conversation, marking the beginning of a new dialogue segment. \\
- preceding\_topical\_relation-YES: The current turn has **some or any** topical/subtopical relation to the preceding conversation context. \\
- intent-INFORMATION SEEKING: The user wants to find factual information or answers to specific questions. \\
\{\textrm{remaining intents and descriptions here}\} \\ 
\text{*}/ \\
\#\# TASK \#\# \\
Using valid SQLite, answer the following multi-turn conversational questions for the table provided above. Use the following steps:  \\
- For each user-agent turn starting with T, output the answer SQL query.  \\
- When preceding\_topical\_relation is YES, you must use the exact same intent and domain label for all turns in the segment.  \\
- Output your answer as a list, with one SQL query per turn starting with T.  \\
\#\# OUTPUT FORMAT \#\# \\
T\{turn number\}. SELECT * from states WHERE preceding\_topical\_relation = \{your answer\} AND intent = \{your\_answer\} AND domain = \{your answer\};  \\
\#\# INPUT \#\# \\
\{\textrm{input dialogue}\} \\
\#\# OUTPUT \#\# \\
}}

\section{Annotation Details}
\label{app:annotation}

\subsection{Labels provided to annotators}
\label{app:annotation-labels}

Below, we provide the labels and descriptions, if available, that were given to the Bing Chat dataset annotators.
For intent and domain, we developed the label names and intent descriptions using an iterative, semi-automated process in which we asked GPT4 to summarize a sample of conversation logs, extract the key themes, and compare these themes to identify the main differences among different types of intents and domains.  

\paragraph{IsSegmentBoundary}
\begin{itemize}
    \setlength{\itemsep}{0.05mm}
    \item NO: The  current turn has no syntactic, semantic, or topical relation to the preceding conversation context OR is the first turn in the conversation. 
    \item YES: The current turn has any syntactic, semantic, or topical relation to the preceding conversation context. 
\end{itemize}

\paragraph{SegmentIntent}
\begin{itemize}
    \setlength{\itemsep}{0.05mm}
    \item INFORMATION SEEKING: The user wants to find factual information or answers to specific questions. 
    \item ANALYSIS: The user asks analytical or conceptual questions about a complex topic or problem. The user's questions require some degree of reasoning, interpretation, argumentation, comparison, and/or data processing.
    \item CREATION: The user asks the agent to either generate original content or translate existing content into new content based on specified criteria or constraints.
    \item OPEN-ENDED DISCOVERY: The user wants to casually chat or play with the agent out of curiosity, boredom, or humor, OR the user's intent is so unclear/underspecified that it's impossible to categorize in any of the other intent classes. The user mainly treats the agent as a conversation or chitchat partner, and none of the other intent categories can be assigned. 
\end{itemize}

\paragraph{SegmentDomain}
\begin{itemize}
\setlength{\itemsep}{0.05mm}
\item AI MACHINE LEARNING AND DATA SCIENCE 
\item ASTROLOGY 
\item BIOLOGY AND LIFE SCIENCE 
\item BUSINESS AND MARKETING 
\item CAREER AND JOB APPLICATION 
\item CLOTHING AND FASHION 
\item COOKING FOOD AND DRINKS 
\item CRAFTS 
\item CULTURE AND HISTORY 
\item CYBERSECURITY 
\item DATING FRIENDSHIPS AND RELATIONSHIPS 
\item DESIGN 
\item EDUCATION 
\item ENTERTAINMENT 
\item ENVIRONMENT AGRICULTURE AND ENERGY 
\item FAMILY PARENTING AND WEDDINGS 
\item FINANCE AND ECONOMICS 
\item GAMES 
\item GEOGRAPHY AND GEOLOGY 
\item HEALTH AND MEDICINE 
\item HOUSING AND HOMES 
\item HUMOR AND SARCASM 
\item LANGUAGE 
\item LAW AND POLITICS 
\item LITERATURE AND POETRY 
\item MANUFACTURING AND MATERIALS 
\item MATH LOGIC AND STATISTICS 
\item MUSIC AND AUDIO 
\item NEWS 
\item PETS AND ANIMALS 
\item PHILOSOPHY 
\item PHYSICS CHEMISTRY AND ASTRONOMY 
\item PRODUCTIVITY 
\item PSYCHOLOGY AND EMOTIONS 
\item RELIGION AND MYTHOLOGY 
\item SHIPPING AND DELIVERY 
\item SHOPPING AND GIFTS 
\item SMALL TALK 
\item SOCIAL MEDIA 
\item SOFTWARE AND WEB DEVELOPMENT 
\item SPORTS AND FITNESS 
\item TAXATION 
\item TECHNOLOGY 
\item TIME AND DATES 
\item TRANSPORTATION AUTOMOTIVE AND AEROSPACE 
\item TRAVEL 
\item VISUAL ARTS AND PHOTOGRAPHY 
\item WEATHER 
\item WRITING JOURNALISM AND PUBLISHING
\end{itemize}

\subsection{Domain labeling procedure}
\label{app:domain}

Due to the large number of domain values and the potential for high disagreement and cognitive overload, we did not ask annotators to choose from the full list of domains per turn.
Rather, we provided a dropdown list of five options per turn.
One option was manually selected by the authors as being correct or near-correct.
Two options were chosen at random using Python.
One option was ``OTHER,'' in which case the annotator was required to choose the correct domain from the full list of 49 domains and explain their choice. 

Finally, the last option was a ``hard negative'' chosen using the following procedure.
First, we manually grouped our domains into eight high-level clusters: STEM, arts, social sciences, health, commerce, professional, personal, and leisure.
Then, given the aforementioned ``ground-truth'' domain chosen by the authors, we randomly sampled another domain from the same high-level cluster as the ground-truth label.
For example, if the ground-truth domain was chosen to be ``BIOLOGY AND LIFE SCIENCE'', we sampled another domain from the STEM cluster as our final domain candidate.